%% file: main.tex
\documentclass[sigconf,natbib=true,anonymous=false,screen=true]{acmart}

\usepackage{lipsum}
\usepackage{multirow}
\usepackage{makecell}
\usepackage{cleveref}
\usepackage{listings}
\usepackage{subcaption}
\usepackage{booktabs}
\usepackage{multirow}
\usepackage{arydshln}
\usepackage[inline]{enumitem}
\usepackage{fontawesome}

\AtBeginDocument{%
  }

\copyrightyear{2026}
\acmYear{2026}
\setcopyright{cc}
\setcctype{by}
\acmConference[SIGIR '26]{Proceedings of the 49th International ACM SIGIR Conference on Research and Development in Information Retrieval}{July 20--24, 2026}{Melbourne, VIC, Australia}
\acmBooktitle{Proceedings of the 49th International ACM SIGIR Conference on Research and Development in Information Retrieval (SIGIR '26), July 20--24, 2026, Melbourne, VIC, Australia}
\acmDOI{10.1145/3805712.3809873}
\acmISBN{979-8-4007-2599-9/2026/07}

\begin{document}

\title{Learning Evidence of Depression Symptoms via Prompt Induction}

\author{Eliseo Bao}
\email{eliseo.bao@udc.es}
\orcid{0009-0000-8457-1115}
\affiliation{
  \institution{IRLab, CITIC, Universidade da Coruña}
  \city{A Coruña}
  \country{Spain}
}

\author{Anxo Perez}
\email{anxo.pvila@udc.es}
\orcid{0000-0002-0480-006X}
\affiliation{%
  \institution{IRLab, CITIC, Universidade da Coruña}
  \city{A Coruña}
  \country{Spain}
}

\author{David Otero}
\email{david.otero.freijeiro@udc.es}
\orcid{0000-0003-1139-0449}
\affiliation{%
  \institution{IRLab, CITIC, Universidade da Coruña}
  \city{A Coruña}
  \country{Spain}
}

\author{Javier Parapar}
\email{javier.parapar@udc.es}
\orcid{0000-0002-5997-8252}
\affiliation{%
  \institution{IRLab, CITIC, Universidade da Coruña}
  \city{A Coruña}
  \country{Spain}
}

\begin{abstract}
Depression places substantial pressure on mental health services, and many people describe their experiences outside clinical settings in high-volume user-generated text (e.g., online forums and social media). Automatically identifying clinical symptom evidence in such text can therefore complement limited clinical capacity and scale to large populations. We address this need through sentence-level classification of 21 depression symptoms from the BDI-II questionnaire, using \textsc{BDI-Sen}, a dataset annotated for symptom relevance. This task is fine-grained and highly imbalanced, and we find that common LLM approaches (zero-shot, in-context learning, and fine-tuning) struggle to apply consistent relevance criteria for most symptoms. We propose Symptom Induction (SI), a novel approach which compresses labeled examples into short, interpretable guidelines that specify what counts as evidence for each symptom and uses these guidelines to condition classification. Across four LLM families and eight models, SI achieves the best overall weighted F1 on \textsc{BDI-Sen}, with especially large gains for infrequent symptoms. Cross-domain evaluation on an external dataset further shows that induced guidelines generalize across other diseases shared symptomatology (bipolar and eating disorders).
\end{abstract}

%
%




\maketitle

\input{sections/1-introduction}
\input{sections/2-methodology}

\input{sections/3-experiments_and_results}

\input{sections/4-conclusion}

\begin{acks}
The first author acknowledges the support of the Department of Education, Science, Universities, and Vocational Training of the Xunta de Galicia (grant ED481A-2024-079). All authors affiliated with IRLab and CITIC acknowledge funding from the Ministry of Science, Innovation and Universities of the Government of Spain (project PID2022-137061OB-C21, MCIN/AEI/10.13039/501100011033), as well as from the Department of Education, Science, Universities, and Vocational Training of the Xunta de Galicia (grant GRC ED431C 2025/49). CITIC, as a center accredited for excellence within the Galician University System and a member of the CIGUS Network, receives subsidies from the Department of Education, Science, Universities, and Vocational Training of the Xunta de Galicia. Additionally, it is co-financed by the EU through the FEDER Galicia 2021-27 operational program (Ref. ED431G 2023/01).
\end{acks}

\bibliographystyle{ACM-Reference-Format}
\bibliography{references}


\end{document}

%% file: sections/1-introduction.tex
\section{Introduction and Background}
\label{sec:introduction}

Public health services have limited capacity to identify and monitor individuals at risk of developing mental health problems~\cite{dgoodwin2022trends,healthorganization2025world}. At the same time, many individuals share their experiences outside clinical settings in social media~\cite{dechoudhury2014mental}. Automatically identifying clinically meaningful symptom evidence in such sources can support large-scale screening research and facilitate clinician-in-the-loop review~\cite{bao2025how}. Recent work increasingly explores using Large Language Models (LLMs) for depression-related classification and analysis~\cite{ge2025survey,lan2025depression,yang2024mentallama}. Nevertheless, binary classification frameworks provide limited insight into the symptom-level evidence underlying predictions, which limits interpretability and clinical utility~\cite{yang2024mentallama,ge2025survey}.

A symptom-grounded perspective can be more actionable: rather than predicting a diagnosis, models search clinically recognized symptoms expressed in text to support assessment and professional review. Prior work has advanced symptom-focused modeling via explainable marker identification with transformers~\cite{bao2024explainable,wang2024explainable}, prompt-learning and graph-based approaches~\cite{chen2025heterogeneous}, LLM-centered methods emphasizing interpretability~\cite{belcastro2025detecting,lan2025depression,yang2024mentallama}, and clinical rationale distillation~\cite{song2025does}. Related efforts connect unstructured text to standardized clinical instruments, such as the BDI-II~\cite{beck1996beck}, leveraging psychometric structure to make screening criteria explicit~\cite{ravenda2025are,rssola2025incidence,bao2025redsm5}.

In this work, we study sentence-level detection of 21 symptoms from the BDI-II questionnaire\footnote{\href{https://github.com/IRLab-UDC/depression-prompt-induction/blob/main/data/BDI21.pdf}{BDI-21 Questionnaire}}~\cite{beck1996beck} using \textsc{BDI-Sen}~\cite{perez2023bdi-sen}, a dataset annotating whether sentences provide first-person symptom evidence. This setting is difficult because
\begin{enumerate*}[label=(\roman*),itemjoin={{, }},itemjoin*={{, and }},afterlabel=~]
\item many symptoms have few labeled examples in the dataset and are linguistically subtle \item semantically related yet non-informative sentences create hard negatives
\end{enumerate*}.
Common LLM strategies for classification include zero-shot prompting, in-context learning (ICL), and supervised fine-tuning (SFT)~\cite{mishra2023prompting,yang2024mentallama,ge2025survey}. However, symptom detection requires compact, stable, symptom-specific decision rules that generalize beyond a handful of examples without long prompts or extensive parameter updates.

We propose \emph{Symptom Induction} (SI), which transforms labeled training instances into concise, interpretable natural-language \emph{guidelines}~\cite{honovich2023instruction,xiao2026prompt-mii} that specify what counts as evidence for each symptom.\footnote{Code available at \faicon{github} \url{https://github.com/IRLab-UDC/depression-prompt-induction}.} Unlike ICL, which uses demonstrations as transient context, SI compresses demonstrations into reusable guidelines, effectively finding symptom guidelines that separates relevant from non-relevant evidence. We compare SI against zero-shot prompting, ICL, and SFT on \textsc{BDI-Sen}~\cite{perez2023bdi-sen}. Across four LLM families and eight models, SI achieves the strongest overall weighted F1, with particularly large gains on infrequent symptoms. To assess robustness, we additionally evaluate cross-domain generalization on \textsc{PsySym}~\cite{zhang2022symptom}, where induced guidelines transfer across different clinical diseases (depression, bipolar disorder, eating disorder) and symptom frameworks. At the same time, we also reflect how all methods perform poorly on extremely rare symptoms, reflecting the limits of scarce data.

Our contributions are: \begin{enumerate*}[label=(\roman*),itemjoin={{; }},itemjoin*={{; and }},afterlabel=~]
\item we introduce Symptom Induction for symptom-level depression classification, framing it as induced symptom-relevance criteria
\item we provide a comparative evaluation of inference strategies (zero-shot, ICL, SFT, SI) on \textsc{BDI-Sen} and \textsc{PsySym}
\item we analyze per-symptom behavior, identifying where SI helps most and where strategies fail.
\end{enumerate*}

%% file: sections/2-methodology.tex
\section{Methodology}
\label{sec:methodology}

We model symptom detection as multi-label binary relevance task: for each sentence, we predict its relevance to each of the 21 BDI-II symptoms. This section describes the task formulation and classification strategies.

\paragraph{\textbf{Task Formulation}}

Given a sentence $s$ and a symptom $c$ from the BDI-II, the goal is to predict whether $s$ contains first-person evidence consistent with symptom $c$. Formally, we learn a function $f:(s,c)\rightarrow\{0,1\}$, where $f(s,c)=1$ indicates relevance and $f(s,c)=0$ otherwise. The BDI-II defines 21 symptoms, and we treat this as 21 independent binary classification tasks under the binary relevance framework.

\subsection{Classification Strategies}

We compare four approaches to symptom classification, each representing a distinct way to condition LLM behavior on the task structure. All methods use the same base LLMs and produce binary predictions for each symptom-sentence pair.

\emph{Zero-Shot Prompting (\textbf{ZS}).}~ZS provides the symptom name and its BDI-II definition, and asks the LLM to judge whether sentence $s$ is relevant to the symptom using a generic instruction. This baseline study the pre-trained LLMs knowledge performance without task-specific adaptation.

\emph{In-Context Learning (\textbf{ICL}).}~ICL extends ZS by adding labeled demonstrations from \textsc{BDISen} to the input context. For each symptom $c$, we include up to 15 positive and 15 negative training examples (fewer for low-frequency symptoms). ICL allows the LLM to infer symptom-specific patterns from the demonstrations at inference time, constructing longer prompts.

\emph{Supervised Fine-Tuning (\textbf{SFT}).}~SFT adapts LLM parameters on the \textsc{BDI-Sen} training data using LoRA~\cite{jhu2022lora}. After finetuning, predictions are made without demonstrations at inference time. SFT learns symptom-specific decision boundaries through gradient updates on the training distribution.

\emph{Symptom Induction (\textbf{SI}).}~We propose \emph{Symptom Induction} (SI), which induces a compact, interpretable guideline for each symptom and then uses it as a reusable decision rule. SI has two stages: (i) an \emph{induction} phase that generates a symptom guideline from labeled examples, and (ii) an \emph{assessment} phase that applies this guideline to classify new sentences. Importantly, SI decouples guideline generation from classification: a fixed \emph{teacher} model generates the guidelines once during induction, and these guidelines can then be reused by different \emph{student} models during assessment, enabling knowledge transfer across model families.

\emph{\textbf{Induction phase.}}~For each symptom $c$, we construct a demonstration set $D_c = \{(s_i, y_i)\}_{i=1}^{N}$ from the training data, where $N = \min(15, |S_c^+|)$ and $S_c^+$ denotes the set of positive training examples for symptom $c$ (we selected 15 based on pilot experiments balancing guideline quality and demonstration set size). We sample an equal number of negative examples from hard negatives (sentences mentioning other symptoms) and soft negatives (control sentences), creating a balanced demonstration set. We then prompt a fixed \emph{induction model} to generate a concise natural-language guideline that describes what counts as evidence for symptom $c$. The choice of $N{=}15$ reflects a practical trade-off: too few examples yield underspecified guidelines, while more examples increase prompt length without proportional quality gains for rare symptoms where positives are inherently scarce. The induced guideline follows a fixed structure designed to capture explicit relevance criteria. It includes a core question testing symptom relevance, patterns specifically indicating symptom presence, patterns that do not qualify as relevant, symptom-specific vocabulary, and edge cases requiring disambiguation. This structured format ensures the guideline addresses both positive indicators and common sources of false positives\footnote{Guidelines in \faicon{github} \url{https://github.com/IRLab-UDC/depression-prompt-induction/blob/main/data/si.json}.}. 

\emph{\textbf{Assessment phase.}}~At inference time, we inject the induced guideline into the system prompt. The classification prompt combines the symptom definition, the induced guideline, and the test sentence, asking the model to apply the guideline's criteria to determine binary relevance. Unlike ICL, which requires the model to interpret demonstrations for every input, SI reuses the same guideline across all test sentences for symptom $c$, yielding shorter inference prompts and a more stable decision criterion. Moreover, the induced guidelines can be reviewed by domain experts, refined based on error analysis, or adapted to new contexts without retraining. Prompt templates for all strategies are available in the repository.

%% file: sections/3-experiments_and_results.tex
\section{Experiments and Results}
\label{sec:experiments_and_results}

We evaluate the four classification strategies on \textsc{BDI-Sen} (in-domain) and \textsc{PsySym} (cross-domain), comparing overall performance, per-symptom behavior, and also generalization across different diseases.

\subsection{Experimental Setup}

\subsubsection{Datasets}

\begin{table}
    \centering
    
    \caption{Dataset statistics. BDI-Sen: training and in-domain evaluation. PsySym: cross-domain evaluation.}
    \label{tab:dataset_statistics}
    \begin{tabular}{lrrrr}
    \toprule
    & \multicolumn{2}{c}{\textbf{BDI-Sen}~\cite{perez2023bdi-sen}} & \multicolumn{2}{c}{\textbf{PsySym}~\cite{zhang2022symptom}} \\
    \cmidrule(lr){2-3} \cmidrule(lr){4-5}
    & \textbf{Train} & \textbf{Test} & \textbf{Train} & \textbf{Test} \\
    \midrule
    Sentences & 762 & 624 & 11{,}220 & 9{,}042 \\
    \quad w/ symptoms & 381 & 104 & 1{,}870 & 1{,}507 \\
    Pos annotations & 631 & 122 & 3{,}908 & 3{,}205 \\
    Disorders & 1 & 1 & 7 & 7 \\
    \bottomrule
    \end{tabular}
\end{table}

We use \textsc{BDI-Sen}~\cite{perez2023bdi-sen} for in-domain evaluation and \textsc{PsySym}~\cite{zhang2022symptom} to test cross-domain generalization. \textsc{BDI-Sen} contains 1{,}516 sentences from Reddit mental health communities annotated for 21 BDI-II symptoms, distinguishing \emph{hard negatives} (other symptoms or past/hypothetical mentions) from \emph{soft negatives} (unrelated content). Table~\ref{tab:dataset_statistics} shows data splits. Symptom frequency varies considerably, posing a central challenge for all strategies. \textsc{PsySym} contains sentence-level symptom annotations from Reddit posts across 7 mental disorders (38 symptom classes), including depression, bipolar disorder, and eating disorder. For cross-domain evaluation, we use a clinician-validated mapping that aligns \textsc{PsySym}'s 14 depression-related symptoms to the 21 BDI-II symptoms based on conceptual overlap. The complete mapping is available in the repository.

\subsubsection{Models}

We evaluate eight LLMs from four families: \textsc{Gemma 3} (4B, 12B), \textsc{Llama 3} (3.2-3B, 3.1-8B), \textsc{Phi-4} (Mini, Base), and \textsc{Qwen3} (4B, 14B). All models run with 4-bit quantization. For SI, we use a fixed \textit{teacher} model, \textsc{Gemma 3 27B}, to induce symptom guidelines from \textsc{BDI-Sen} training examples. The same induced guidelines are then applied to each \textit{student} model at inference time.

\subsubsection{Implementation Details}

We use vLLM for efficient batch inference with structured output constraints, restricting model responses to ``YES''/``NO'' tokens. For ZS and ICL, we apply the base instruction-tuned models directly. For SFT, we fine-tune each base model using LoRa (specific hyperparameters are available in the repository). For SI, we use \textsc{Gemma 3 27B} to generate guidelines from balanced demonstration sets (up to 15 positive and 15 negative examples per symptom), then apply the resulting guidelines at inference with the base models using a fixed prompt template.

\subsection{Evaluation}

We report (i) per-symptom F1 and (ii) an overall weighted F1 across all symptoms. We use weighted F1 rather than macro or micro F1 because it accounts for symptom frequency by weighting each symptom's F1 by its number of positive test samples, balancing the influence of rare and frequent symptoms while remaining interpretable at the symptom level. We also report the unweighted mean F1 across symptoms (equivalent to macro F1) in the per-symptom tables for comparability with prior work on \textsc{BDI-Sen}~\cite{perez2023bdi-sen}.

\subsection{Overall Performance Comparison}

\begin{table}
    \centering
    \caption{Weighted F1 on BDI-Sen. Best figure per row is bolfaced, while second best is \underline{underlined}.}
    \label{tab:bdi_sen_overall_f1}
    \begin{tabular}{lcccc}
    \toprule
    \textbf{Model} & \textbf{ZS} & \textbf{ICL} & \textbf{FT} & \textbf{SI} \\
    \midrule
    \textsc{Gemma 3 4B} & \underline{0.267} & 0.171 & 0.242 & \textbf{0.389} \\
    \textsc{Gemma 3 12B} & \underline{0.296} & 0.268 & 0.282 & \textbf{0.367} \\
    \textsc{Llama 3.2 3B} & \textbf{0.230} & 0.138 & 0.203 & \underline{0.229} \\
    \textsc{Llama 3.1 8B} & \underline{0.227} & 0.200 & 0.223 & \textbf{0.275} \\
    \textsc{Phi-4 Mini} & \underline{0.277} & 0.167 & \textbf{0.311} & \underline{0.277} \\
    \textsc{Phi-4} & 0.279 & 0.196 & \underline{0.305} & \textbf{0.329} \\
    \textsc{Qwen3 4B} & 0.285 & 0.267 & \underline{0.299} & \textbf{0.359} \\
    \textsc{Qwen3 14B} & \underline{0.296} & 0.199 & 0.263 & \textbf{0.348} \\
    \bottomrule
    \end{tabular}
\end{table}

\begin{figure}
    \centering
    \includegraphics[width=0.60\linewidth]{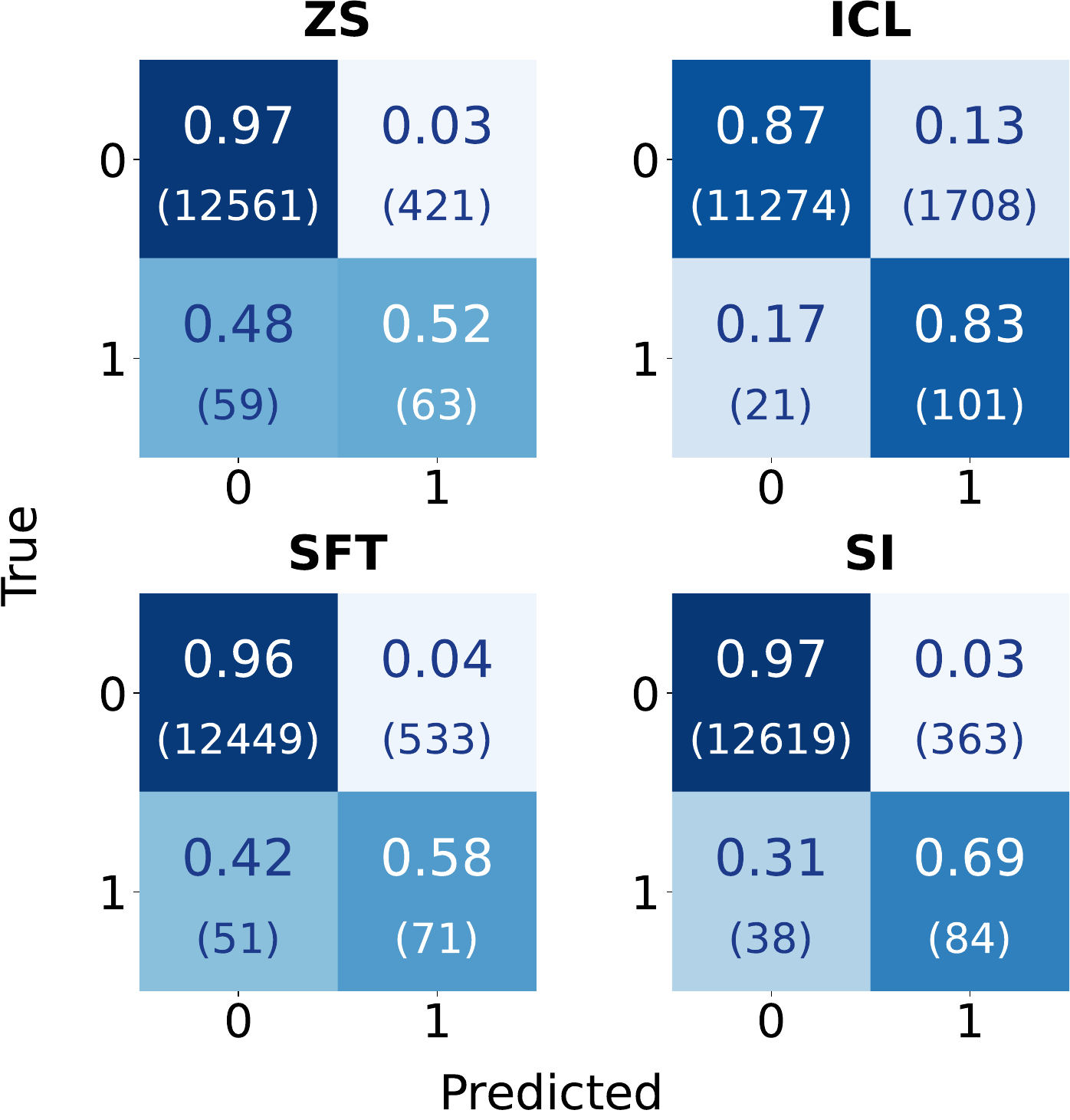}
    \caption{Aggregated confusion matrices for \textsc{Gemma 3 4B} on BDI-Sen. Row-normalized, aggregated across all symptoms.}
    \Description{Confusion matrices showing the distribution of true positives, false positives, true negatives, and false negatives for four classification strategies (ZS, ICL, SFT, SI) on BDI-Sen.}
    \label{fig:overall_matrices}
\end{figure}

Table~\ref{tab:bdi_sen_overall_f1} reports weighted F1 on \textsc{BDI-Sen} across models and  strategies. SI achieves the best score for six of the eight models and is competitive on the remaining two. The strongest configuration is \textsc{Gemma 3 4B} with SI (0.389), improving substantially over ZS (0.267), ICL (0.171), and SFT (0.242). Notably, strategy choice can dominate model scale: \textsc{Gemma 3 4B} with SI outperforms all other model/strategy combinations in our study, including larger models with fine-tuning. Figure~\ref{fig:overall_matrices} helps explain these differences for \textsc{Gemma 3 4B} by showing its aggregated confusion matrices across all symptom--sentence pairs. ICL exhibits a clear over-prediction tendency, producing many more false positives than the other strategies (1708 FPs), while missing very few positives (21 FNs). In contrast, ZS and SFT are more conservative: both reduce false positives (421 and 533 FPs) but at the cost of higher false negatives (5 and 51 FNs), indicating that they miss substantial symptom evidence. SI achieves the best balance, reducing false positives relative to ZS/SFT while also lowering false negatives (38 FNs).

These patterns suggest distinct failure modes. ($i$) ICL is sensitive to data sparsity and demonstration quality: for low-frequency symptoms, the available positives are limited and heterogeneous, and the model may use superficial cues. Moreover, mixing hard negatives in the context can be beneficial in principle, but in practice it can confuse the model when positives and negatives differ only subtly. ($ii$) Fine-tuning shows mixed results: it improves over ZS for some models (e.g., \textsc{Phi-4 Mini}) but is less reliable under strong imbalance, where parameter updates can overfit or fail to learn robust symptom-specific criteria. ($iii$) Finally, SI mitigates both issues by distilling demonstrations into an explicit, interpretable guideline: its relevance criteria directly encode inclusion and exclusion patterns, helping models reject hard negatives without requiring few-shot prompts or parameter updates.

\subsection{Per-Symptom Analysis}

\begin{table}
    \centering
    \caption{Per-symptom F1 on BDI-Sen for \textsc{Gemma 3 4B}. Best figure per row is bolfaced, while second best is \underline{underlined}}
    \label{tab:bdisen_per_symptom_f1}
    \begin{tabular}{lcccc}
        \toprule
        \textbf{Symptom} & \textbf{ZS} & \textbf{ICL} & \textbf{FT} & \textbf{SI} \\
        \midrule
        Sadness                        & 0.455 & 0.309 & \underline{0.463} & \textbf{0.469} \\
        Self-Dislike                   & \underline{0.123} & 0.102 & 0.105 & \textbf{0.227} \\
        Loss of Pleasure               & 0.154 & \underline{0.224} & 0.120 & \textbf{0.508} \\
        Past Failure                   & \underline{0.231} & 0.137 & 0.176 & \textbf{0.356} \\
        Pessimism                      & \textbf{0.320} & 0.092 & 0.262 & \underline{0.318} \\
        Irritability                   & \underline{0.255} & 0.096 & 0.188 & \textbf{0.286} \\
        Worthlessness                  & 0.043 & 0.045 & \underline{0.049} & \textbf{0.080} \\
        Guilty Feelings                & 0.059 & 0.032 & \textbf{0.118} & \underline{0.071} \\
        Tiredness or Fatigue           & \textbf{0.462} & 0.162 & 0.333 & \underline{0.353} \\
        Crying                         & \underline{0.556} & 0.088 & 0.345 & \textbf{0.714} \\
        Loss of Interest               & 0.000 & 0.056 & \underline{0.095} & \textbf{0.286} \\
        Suicidal Thoughts              & 0.318 & 0.238 & \underline{0.341} & \textbf{0.500} \\
        Punishment Feelings            & 0.000 & \underline{0.021} & 0.000 & \textbf{0.129} \\
        Indecisiveness                 & \underline{0.333} & 0.125 & \underline{0.333} & \textbf{0.500} \\
        Concentration Difficulty       & \underline{0.222} & 0.040 & 0.167 & \textbf{0.500} \\
        Loss of Energy                 & \underline{0.000} & \textbf{0.051} & \underline{0.000} & \underline{0.000} \\
        Agitation                      & 0.100 & 0.056 & \textbf{0.154} & \underline{0.105} \\
        Changes in Appetite            & \textbf{0.500} & 0.182 & \textbf{0.500} & \textbf{0.500} \\
        Changes in Sleeping Pattern    & 0.333 & 0.235 & \textbf{0.500} & \underline{0.400} \\
        Self-Criticalness              & 0.031 & 0.016 & \underline{0.034} & \textbf{0.045} \\
        Loss of Interest in Sex        & \textbf{0.000} & \textbf{0.000} & \textbf{0.000} & \textbf{0.000} \\
        \midrule
        \textit{Raw average} & \underline{0.214} & 0.110 & 0.204 & \textbf{0.302} \\
        \bottomrule
    \end{tabular}
\end{table}

\begin{table*}
    \centering
    \caption{Per-symptom F1 on PsySym for \textsc{Gemma 3 4B}. Best figure per row is bolfaced, while second best is \underline{underlined}. A ``-'' indicates a symptom not present.}
    \label{tab:psysym_per_symptom}
    \begin{tabular}{lcccc:cccc:cccc}
        \toprule
        \multirow{2}{*}[-2pt]{\textbf{Symptom}} & \multicolumn{4}{c}{\textbf{Depression}} & \multicolumn{4}{c}{\textbf{Bipolar Disorder}} & \multicolumn{4}{c}{\textbf{Eating Disorder}} \\
        \cmidrule(lr){2-5}\cmidrule(lr){6-9}\cmidrule(lr){10-13}
         & \textbf{ZS} & \textbf{ICL} & \textbf{SFT} & \textbf{SI} & \textbf{ZS} & \textbf{ICL} & \textbf{SFT} & \textbf{SI} & \textbf{ZS} & \textbf{ICL} & \textbf{SFT} & \textbf{SI} \\
        \midrule
        Sadness                        & \underline{0.535} & 0.493 & \textbf{0.540} & 0.532 & 0.394 & \underline{0.411} & \textbf{0.477} & 0.377 & \underline{0.424} & 0.396 & 0.400 & \textbf{0.571} \\
        Self-Dislike                   & \underline{0.200} & 0.186 & 0.179 & \textbf{0.430} & \underline{0.571} & 0.227 & \textbf{0.615} & 0.561 & \underline{0.387} & 0.320 & \textbf{0.406} & 0.385 \\
        Loss of Pleasure               & \underline{0.636} & 0.245 & 0.550 & \textbf{0.695} & \textbf{0.412} & 0.170 & \underline{0.364} & 0.357 & -- & -- & -- & -- \\
        Past Failure                   & 0.372 & 0.242 & \underline{0.388} & \textbf{0.466} & 0.267 & 0.284 & \underline{0.294} & \textbf{0.528} & 0.056 & \textbf{0.378} & 0.154 & \underline{0.314} \\
        Pessimism                      & \textbf{0.808} & 0.564 & \underline{0.762} & 0.731 & -- & -- & -- & -- & -- & -- & -- & -- \\
        Irritability                   & \underline{0.698} & 0.230 & 0.500 & \textbf{0.757} & \textbf{0.840} & 0.430 & \underline{0.694} & 0.667 & \underline{0.483} & 0.299 & 0.421 & \textbf{0.560} \\
        Worthlessness                  & \underline{0.396} & 0.303 & \textbf{0.400} & 0.308 & 0.564 & 0.416 & \textbf{0.615} & \underline{0.585} & 0.056 & \textbf{0.417} & 0.054 & \underline{0.205} \\
        Guilty Feelings                & \underline{0.568} & 0.299 & 0.518 & \textbf{0.628} & 0.235 & \textbf{0.454} & 0.400 & \underline{0.450} & 0.098 & \textbf{0.379} & 0.130 & \underline{0.273} \\
        Tiredness or Fatigue           & 0.778 & 0.585 & \textbf{0.785} & \underline{0.779} & 0.450 & 0.444 & \underline{0.531} & \textbf{0.698} & -- & -- & -- & -- \\
        Crying                         & 0.133 & \textbf{0.475} & \underline{0.224} & 0.083 & 0.042 & \textbf{0.376} & \underline{0.120} & 0.000 & 0.333 & \underline{0.451} & \textbf{0.457} & 0.345 \\
        Loss of Interest               & \textbf{0.735} & 0.330 & \underline{0.652} & 0.580 & \underline{0.429} & 0.265 & 0.412 & \textbf{0.529} & -- & -- & -- & -- \\
        Suicidal Thoughts              & \textbf{0.747} & 0.581 & 0.688 & \underline{0.724} & 0.508 & \textbf{0.764} & \underline{0.649} & 0.385 & -- & -- & -- & -- \\
        Punishment Feelings            & 0.105 & \underline{0.270} & 0.257 & \textbf{0.619} & 0.071 & \underline{0.324} & 0.114 & \textbf{0.478} & 0.100 & \underline{0.365} & 0.089 & \textbf{0.392} \\
        Indecisiveness                 & \textbf{0.804} & 0.717 & \underline{0.800} & 0.637 & -- & -- & -- & -- & -- & -- & -- & -- \\
        Concentration Difficulty       & \underline{0.643} & 0.619 & \textbf{0.728} & 0.581 & \underline{0.778} & 0.370 & 0.758 & \textbf{0.784} & -- & -- & -- & -- \\
        Loss of Energy                 & \textbf{0.779} & 0.318 & 0.635 & \underline{0.706} & 0.489 & 0.392 & \underline{0.510} & \textbf{0.560} & -- & -- & -- & -- \\
        Agitation                      & \textbf{0.677} & 0.441 & 0.560 & \underline{0.581} & \textbf{0.797} & 0.549 & \underline{0.741} & 0.681 & \underline{0.519} & 0.329 & 0.375 & \textbf{0.526} \\
        Changes in Appetite            & \textbf{0.899} & 0.857 & \underline{0.873} & 0.767 & \textbf{0.875} & \underline{0.800} & 0.757 & 0.667 & 0.790 & \textbf{0.882} & \underline{0.839} & 0.623 \\
        Changes in Sleeping Pattern    & \underline{0.750} & 0.309 & \textbf{0.844} & 0.667 & \underline{0.727} & 0.500 & \textbf{0.760} & 0.632 & \underline{0.857} & 0.216 & \textbf{1.000} & 0.667 \\
        Self-Criticalness              & \underline{0.440} & 0.202 & 0.430 & \textbf{0.481} & \textbf{0.578} & 0.239 & \underline{0.500} & 0.462 & 0.208 & \underline{0.354} & 0.250 & \textbf{0.370} \\
        Loss of Interest in Sex        & \underline{0.765} & 0.724 & 0.743 & \textbf{0.800} & -- & -- & -- & -- & 0.250 & \textbf{0.571} & \underline{0.444} & 0.250 \\
        \midrule
        \textit{Raw average} & \underline{0.594} & 0.428 & 0.574 & \textbf{0.598} & 0.501 & 0.412 & \underline{0.517} & \textbf{0.522} & 0.351 & \underline{0.412} & 0.386 & \textbf{0.422} \\
        \bottomrule
    \end{tabular}
\end{table*}

We analyze per-symptom behavior using \textsc{Gemma 3 4B}, which achieves the best overall weighted F1, making it a representative setting to study at symptom-level. Table~\ref{tab:bdisen_per_symptom_f1} reports per-symptom F1 on \textsc{BDI-Sen}. SI attains the highest F1 for most symptoms, yielding the best raw average (0.302). Gains are most pronounced for symptoms where evidence is subtle or easily confounded with hard negatives. For example, SI improves \emph{loss of pleasure} (0.508 vs.\ 0.224 ICL), or \emph{loss of interest} (0.286 vs.\ 0.095 SFT). These results suggest that induced guidelines help models apply more consistent symptom-specific criteria rather than relying on a small set of demonstrations.

A qualitative inspection of the induced guidelines helps explain these differences. High-performing guidelines (e.g., \emph{crying}: SI F1 = 0.714) specify precise inclusion criteria and distinctive vocabulary that does not overlap with other symptoms, giving the model a stable decision boundary. Lower-performing ones (e.g., \emph{loss of energy}: SI F1 = 0.000) rely on vocabulary (\textit{tired, fatigue, drained}) that largely overlaps with a closely related symptom (\emph{tiredness or fatigue}), making sentence-level disambiguation difficult regardless of guideline quality.

SI does not dominate uniformly. ZS is strongest for \emph{pessimism} (0.320), suggesting that symptom definition alone can be sufficient for some cases. SFT performs best on \emph{guilty feelings} and \emph{agitation}, indicating that some symptoms benefit from parameter updates that capture distributional patterns in the training data. Several symptoms show comparable performance across strategies (e.g., \emph{changes in appetite}), implying that they may be easier to detect given the available examples. Finally, all approaches struggle on extremely sparse symptoms. \emph{Loss of interest in sex} remains at 0.0 F1 (only one positive test instance), and \emph{loss of energy} reaches at most 0.051, highlighting a fundamental limitation of all approaches when training data is minimal. For these symptoms, the bottleneck is not the classification strategy but the scarcity of labeled evidence: SI cannot induce meaningful criteria from one or two positives, SFT cannot learn a reliable decision boundary, and ICL has no representative examples to draw from. Addressing this would require targeted data augmentation or additional expert annotation for underrepresented symptoms rather than improvements to the classification strategy itself.

\vspace*{-0.6em}

\subsection{Cross-Disease Generalization}

Table~\ref{tab:psysym_per_symptom} reports per-symptom F1 on \textsc{PsySym} using \textsc{Gemma 3 4B}, for depression, bipolar, and eating disorders. To enable this cross-disease evaluation, we align \textsc{PsySym}'s symptom taxonomy to the 21 BDI-II symptoms using a clinician-validated mapping, retaining only symptoms with clear conceptual overlap. This tests whether models trained to recognize BDI-II symptom evidence transfer to other disorders with overlapping symptom manifestations. Looking at the overall results, SI achieves the highest raw average F1 in all three disorders, indicating that induced guidelines remain effective under domain shift.

Compared to in-domain results on \textsc{BDI-Sen}, the relative ordering of strategies shifts. Zero-shot prompting is competitive for depression on \textsc{PsySym} (0.594 vs.\ 0.598 for SI) and is best for several symptoms (e.g., \emph{indecisiveness}, \emph{suicidal thoughts}), suggesting that definitions alone can be strong cues for some symptoms. However, SI retains clear advantages on symptoms where evidence can be easily confounded, such as \emph{punishment feelings} and \emph{past failure}, where explicit criteria better separate evidence. SFT shows mixed transfer, performing well on some symptoms (e.g., \emph{sadness}) but degrading on others (e.g., \emph{self-dislike}), indicating reduced robustness when updates are learned from a small, imbalanced source dataset. ICL remains weakest across disorders (0.412--0.428 raw average), and the SI--ICL gap widens under domain shift, supporting that reusable induced guidelines transfer more reliably than re-interpreting few-shot examples for each input.

%% file: sections/4-conclusion.tex
\section{Conclusion}
\label{sec:conclusion}

We introduced Symptom Induction (SI), a method where a induction model distills labeled examples into concise natural-language guidelines for depression symptom classification. Across four LLM families on \textsc{BDI-Sen}, SI achieves the strongest overall weighted F1, with especially large gains on low-resource symptoms. Compared to supervised fine-tuning, SI yields explicit, interpretable decision criteria, avoids additional training, and is less prone to overfitting in low-data scenarios. One limitation of the current comparison is that SI relies on a larger teacher model (27B) than the student models evaluated (3B--14B), which means some of SI's advantage over SFT may reflect the teacher's capacity rather than the induction paradigm alone; future work should ablate teacher size to disentangle these factors. Beyond this, future work should explore clinician-in-the-loop refinement of induced guidelines, more robust induction objectives, and targeted data augmentation for sparsely observed symptoms. Overall, SI offers a practical alternative for symptom detection that balances interpretability with sample efficiency, supporting scalable mental health screening from user-generated text.